# Scaling Up Cartesian Genetic Programming through Preferential Selection of Larger Solutions

**Nicola Milano and Stefano Nolfi**
Institute of Cognitive Sciences and Technologies
National Research Council (CNR)
Via S. Martino della Battaglia, 44
00185 Roma, Italia
stefano.nolfi@istc.cnr.it, nicola.milano@istc.cnr.it

**Abstract**

*We demonstrate how efficiency of Cartesian Genetic Programming method can be scaled up through the preferential selection of phenotypically larger solutions, i.e. through the preferential selection of larger solutions among equally good solutions. The advantage of the preferential selection of larger solutions is validated on the six, seven and eight-bit parity problems, on a dynamically varying problem involving the classification of binary patterns, and on the "Paige" regression problem. In all cases, the preferential selection of larger solutions provides an advantage in term of the performance of the evolved solutions and in term of speed, the number of evaluations required to evolve optimal or high-quality solutions. The advantage provided by the preferential selection of larger solutions can be further extended by self-adapting the mutation rate through the one-fifth success rule. Finally, for problems like the Paige regression in which neutrality plays a minor role, the advantage of the preferential selection of larger solutions can be extended by preferring larger solutions also among quasi-neutral alternative candidate solutions, i.e. solutions achieving slightly different performance.*

## 1. Introduction

Evolutionary algorithms are known for their ability to discover simple solutions, the tendency to select the smallest available solutions. This is not surprising since evolution typically proceeds by selecting minimal sub-optimal solutions first and by then progressively extending those solutions until an optimal solution is found. From an application point of view, the tendency to select compact solution is also desirable since it permits to generate solutions that are cheaper, lighter, and/or faster than non-compact solutions.

Compact solutions, however, can be less evolvable than more elaborated solutions. Consequently, the tendency to select minimal solutions might reduce the chance to produce behavioral variation as a result of genetic variations --- a property that constitutes a crucial prerequisite for evolutionary progress.

This effect has been shown by the authors in a recent paper involving the evolution of digital circuits selected for the ability to solve a five-bit parity problem (Milano, Pagliuca and Nolfi, 2017). To produce adaptive heritable variations, evolving programs should possess both robustness to genetic variation and phenotypic variability. Robustness is required to enable a reliable transfer of capacities from parents to offspring in the presence of mutations. Phenotypic variability, namely the ability to produce phenotypic variations as a result of genetic variation, is required to generate adaptations. One simple way to achieve robustness consists in minimizing the size of the current candidate solutions since this permits to minimize the chance that mutations alter functional components of the solution and consequently the fitness of the solution. Improving robustness in this way however only provides an advantage on the short term since it causes a reduction phenotypic variability that, in turn, causes a reduction of the chance to produce adaptive variations.

In this paper we verify whether the preferential selection of phenotypically large solutions enables to preserve and/or enhance the evolvability of candidate solutions and consequently the probability to evolve high performing solutions. To achieve this objective, we propose a new Cartesian Genetic Algorithm that preferentially chose larger solutions among alternative solutions having the same fitness. The results collected on six, seven and eight-bit parity problems and on a continuous regression problem indicate that the new algorithm largely outperform the standard algorithm in term of speed and quality of the evolved solutions. Moreover, the results collected on a dynamically varying problem indicate that the preferential selection of larger solutions strongly reduce the time required to adapt to problem variations. We show that the advantage of the preferential selection of larger solutions can be extended by self-adapting the mutation rate through the one-fifth success rule. Finally, we show that for problems in which neutrality plays a minor role, the advantage of the preferential selection of larger solutions can be extended by preferring larger solutions among quasi-neutral alternative candidate solutions, i.e. also among solutions achieving slightly different performance

The paper is organized as follow. In Section 2 we briefly review Cartesian Genetic Programming. In Section 3 we discuss previous evidences concerning the relation between the size and the evolvability of candidate solutions. In section 4 and 5 we describe the experimental method and the obtained results. Finally, in section 6 we draw our conclusions.

**2. Cartesian Genetic Programming**

Cartesian Genetic Programming (CGP, Miller and Thomson, 2000; Miller, 2011) is a form of Genetic Programming (Koza, 1992; Langdon and Poli, 2002) which typically is used to evolve acyclic computational structures of nodes (graph) indexed by their Cartesian coordinates but which can also be extended to evolve recurrent (cyclic) structures (Turner and Miller, 2014). The method has been successfully applied to evolve digital circuits (Miller, Job and Vassilev, 2000a-b), robots' controllers (Harding and Miller, 2005), Atari games players (Wilson, Cussat-Blanc, Luga and Miller, 2018), neural networks (Khan, Ahmad, Khan and Miller, 2013), image classifier (Harding, Graziano, Leitner and Schmidhuber, 2012), molecular docking (Garmendia-Doval, Morley and Juhos, 2003), and regression programs (Harding, Miller and Banzhaf, 2009).

CGP use a vector of integer to encode a graph constituted by nodes and connections among nodes, inputs and outputs. This gives it great generality since it can represent digital circuits, programs, neural networks, or other computation structures. The properties of each node are encoded in a tuple of integers (genes), with one gene specifying the function of the node chosen from a list of available functions (defined by the user) and the remaining genes expressing the indexes of the node's inputs. Nodes can take input from either problem inputs or any node preceding them in the chromosome. A final set of one-gene tuples specify the index of the nodes that are used as outputs (Figure 1).

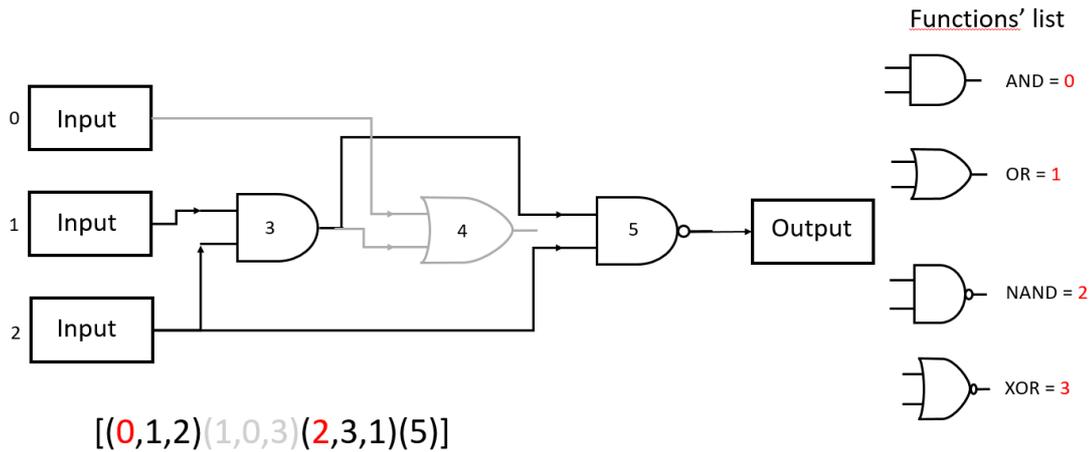

Figure 1. Left: example of a chromosome and of a corresponding CGP program with three inputs, three nodes, and one output. The three-numbers tuples encode the function and the input of the three nodes. The one-number tuples encode the index of the node that are used as outputs. The functional and non-functional genes/nodes are shown in black and grey, respectively. The list of functions shown on the right side show the functions that can be played by nodes and the associated index. Indexes are constituted by integer numbers and are used to identify the individual inputs and then the individual nodes in the order in which they are encoded in the chromosome.

The number of inputs and outputs are fixed and determined on the basis of the nature of the problem. For example, in the case of a six-bits parity problem the number of inputs can be set to 6 and the number of output to 1. The number of gates and the functions that gates can assume is also determined manually by the experimenter on the basis of the nature of the problem. For example, the binary operators shown in Figure 1 are a suitable choice for a parity problem while a list of mathematical operators such as addition, subtraction, multiplication and division, can be a suitable choice for a continuous control problem, like pole-balancing, in which the input and output states are continuous (Tuner and Miller, 2015).

The initial chromosome is generated randomly. More specifically the value of the genes encoding the function of each node are generated randomly with a uniform distribution in the range [0, NGF] where NGF is the number of alternative functions included in the functions' list. The value of the genes encoding the input of a node are generated randomly with a uniform distribution in the range [0, G-1], where G is the index of the corresponding node. Offspring are generated by creating a mutated copy of the parent chromosome. Mutations are realized by replacing a fractions of randomly selected genes with new integer selected randomly with a uniform distribution in the ranges described above. Nodes can eventually be arranged in layers by limiting the range of the input indexes to the nodes contained in the previous layer.

Usually only a small fraction of the nodes participates to the generation of the output values. Indeed, although each node has a function and has inputs, the output of a node does not necessarily have to be used by the inputs of later nodes. The nodes with un-used outputs are non-functional since they do not influence the overall output of the program. Similarly, genes encoding nodes with un-used outputs are non-functional since they do not affect the performance of the program (functional and non-functional nodes and genes are indicated in black and gray in Figure 1). Non-functional nodes can be eliminated from a program without altering its behavior. However, non-functional genes cannot be regarded as noncoding genes since, during evolution, mutations in "downstream" genes (i.e., on the right) can turn them into functional genes. Similarly, formerly functional genes can become non-functional as a result of mutations in other genes (Miller and Smith, 2006).

CGP programs are usually evolved through a $(\mu + \lambda)$ Evolutionary Strategy (Rechenberg, 1973; Beyer and Schwefel, 2002), where $\mu$ is set to 1 and $\lambda$ is usually set to 4 (Miller and Thomson, 2000; Miller, 2011). This means that during each generation a single parent produces four offspring with mutations, as described above. The mutation probability is usually uniform for all genes. When none of the offspring is better than the parent and at least an offspring equals the parent, the offspring is selected. The comparative study reported in Milano, Pagliuca and Nolfi (2017) indicates that the utilization of a single parent might play an important role in CGP. Indeed, CPG programs evolved on the five-bits parity task with $\mu = 1$ largely outperformed programs evolved with $\mu > 1$.

**3. On the relation between size and phenotypic variability**

We now discuss the relation between functional size and phenotypic variability, i.e. the propensity of candidate solutions to vary phenotypically as a result of mutations. The phenotypic variability of a candidate solution can be estimated by measuring the number of programs located in the genetic neighborhood of the original candidate solution displaying different unique behaviors. In the context of digital circuits, programs displaying different unique behaviors correspond to candidate solutions that produce output matrixes that differ among themselves for at least one output value. Phenotypic variability constitutes a crucial prerequisite for adaptive progress. Clearly, in the majority of cases offspring producing outputs that differ from their parent will have a lower fitness than the parent. However, larger the number of different unique solutions located around the candidate solution is, greater the probability that the genetic neighborhood includes at least one adapted solution is.

A positive correlation between size and phenotypic variability was reported by Raman and Wagner (2011). They analyzed digital circuits with randomly generated genotypes that included four inputs, four outputs, and a variable number of nodes. The function list of the nodes included the OR, AND, XOR, NAND and NOR functions. Phenotypic variability was estimated by performing 2,000 steps function-preserving random walk from the original circuit. The random walk was realized by: (i) generating a varied circuit by mutating one gene, (ii) incrementing a counter if the varied circuit produce a different unique behavior (i.e. computes a function that differ from the functions computed by the original circuit and by the previous varied circuits), (iii) preserving or removing the mutation depending on whether the varied circuit has the same fitness of the original circuit or not, (iv) repeating the previous three operations for 2'000 steps. Raman and Wagner (2011) refer to this measure with the term evolvability rather than with the term phenotypic variability. We prefer to distinguish between the propensity of a program to vary phenotypically independently of whether the variations are adaptive or nor (phenotypic variability) and the propensity of a program to generate adaptive variations (evolvability).

The presence of a positive correlation was confirmed by the analysis of digital circuits evolved for the ability to solve a five-bits even parity problem (Milano, Pagliuca and Nolfi, 2018). Indeed, the fitness and the functional size of evolved circuits were strongly correlated. Moreover, the circuits evolved with the $(1 + \lambda)$ ES were significantly larger than the circuits evolved with the $(\mu + \mu)$ ES. This combined with the observation that $(1 + \lambda)$ ES method largely outperform the $(\mu + \mu)$ ES method, with respect to speed and fraction of experiments achieving optimal performance, indicates that the advantage of the $(1 + \lambda)$ ES method is due to the fact that it tends to evolve larger programs with respect to the $(\mu + \mu)$ ES method (Milano, Pagliuca and Nolfi, 2018).

The relation between robustness to mutation and phenotypic variability can be schematized by relying on the notion of neutral network (Shuster et al., 1994; Van Nimwegen, Crutchfield and Huynen, 1999). A neutral network is constituted by a series of nodes connected through bidirectional links in which nodes correspond to candidate solutions with identical fitness and links correspond to single mutations that enable to transform a candidate solution in a genetically different candidate solution achieving the same fitness. The number of links per nodes and consequently the robustness to mutation is higher in the central part of the neutral network (Shuster et al., 1994; Van Nimwegen,

Crutchfield and Huynen, 1999). Under neutral evolution, evolving individuals do not move in an entirely random way over the neutral network but move preferentially in highly connected parts of the network, resulting in phenotypes that are relatively robust against mutations (Van Nimwegen, Crutchfield and Huynen, 1999). Such robust regions of the neutral networks include minimal solutions that achieve robustness by minimizing the number of functional genes but which have a low phenotypic variability (Wagner 2008, 2011; Hu et al., 2012). The preferential selection of larger solutions drives the evolutionary search toward the peripheral areas of the neutral network and/or toward solutions that achieve robustness through redundancy or degeneracy (Tononi, Sporns and Edelman, 1999; Edelman and Gally, 2001, Milano and Nolfi, 2016). These areas are characterized by a greater phenotypical variability than areas including minimal solutions. In other words, the preferential selection of larger solutions drives the evolutionary search toward solutions that are more evolvable.

## 4. Method

As mentioned above, we compared the performance of two algorithms: a standard $(1 + \lambda)$ ES and a $(1 + \lambda)$ ES-PL algorithm that preferentially select program with functionally larger phenotypes. In both cases, during each generation a single parent produces $\lambda$ offspring with mutations. In the case of the $(1 + \lambda)$ ES method, the $1 + \lambda$ candidate solutions are sorted primarily on the basis of their fitness (in descending order), and secondarily on the basis of whether they correspond to an offspring or to the parent (in this order). The first individual is then used to replace the parent. In the case of the $(1 + \lambda)$ ES-PL method, instead, the $(1 + \lambda)$ individuals are sorted primarily on the basis of their fitness (in descending order), secondarily on the basis of the size of their functional circuit (in descending order), and thirdly on the basis of whether they correspond to an offspring or to the parent (in this order). The first individual is then used to replace the parent.

The chromosomes of candidate solutions include 100 three-digit tuples that encode the function and the input of 100 corresponding nodes and N one-digit tuples that encode the index of N corresponding output nodes. The value of the genes encoding the input of a node are generated randomly with an uniform distribution in the range [0, G-1], where G is the index of the corresponding node. Offspring are generated by creating a mutated copy of the parent chromosome. The genes of the initial chromosome are created randomly with a uniform distribution in the range described above. Mutations are realized by replacing a fractions of randomly selected genes with new integers selected randomly with a uniform distribution in the ranges described above.

In the case of the parity and dynamical varying problem, the function list of the nodes included the following four logic operators [AND, NAND, OR, and NOR] and the fitness of a candidate solution was computed on the basis of the inverse of the offset between the outputs produced by the program and the desired outputs. More specifically, the fitness is calculated on the basis of the following equation:

$$F = 1 - \frac{1}{2^n} \sum_{j=1}^{2^n} |O_j - E_j| \qquad (1)$$

where *n* is the number of inputs of the circuit, *j* is the number of the input patterns varying in the range $[1, 2^n]$, $O_j$ is the output of the circuit for pattern *j*, $E_j$ is the desired output for pattern *j*.

In the case of the experiments on the parity problems, programs were evolved for the ability to solve a 6-bit, 7-bit or 8bit even parity problems and were provided with 6, 7, or 8 inputs and 1 output. Programs were evaluated on $2^6$, $2^7$, or $2^8$ patterns, respectively, corresponding to all possible input patterns. The $\lambda$ and the mutrate parameters were set to 4 and 2%. As reported in Section 5.1, these values correspond to the best combination of parameters for the $(1 + \lambda)$ ES method in the case of the 6-bt parity problem. The evolutionary process was terminated when a candidate solution achieved optimal performance or after a total of 1 million candidate solutions were evaluated.

In the case of the dynamically varying problem, programs were provided with 5 inputs and 1 output and were evaluated on $2^5 = 32$ patterns corresponding to all possible input patterns. The problem consists in categorizing the 32 binary input vectors made of 5 binary digits in two categories that correspond to the two possible values that the output node can assume. The category of each input patters is generated randomly at the beginning of each replication with a uniform distribution. Problem variations were realized by switching the desired outputs of a fraction of the patterns selected randomly. The experiments have been replicated by switching the output of 2, 4, 6, 8, 10 or 16 patterns every 100'000 generations.

In the case of the regression problem, we used the Pagie function (2) that constitute a challenging benchmark (Pagie and Hogeweg, 1997; Tuner and Miller, 2015). Following Tuner and Miller (2015), programs are provided with two inputs and 1 output and 100 nodes, the input patters consist of 676 samples taken randomly from the range $x_1$ in [-5.0, 5.0] and $x_2$ in [-5.0, 5.0] with an uniform random distribution, and the functions' set contains the following operators [: +, −, ∗, and / operators]. The fitness function is defined to be the sum of the absolute errors between the output produced on the basis of equation (2) and the value produced by the evolved symbolic program.

$$f(x_1, x_2) = \frac{1}{1 + x_1^{-4}} + \frac{1}{1 + x_2^{-4}} \tag{2}$$

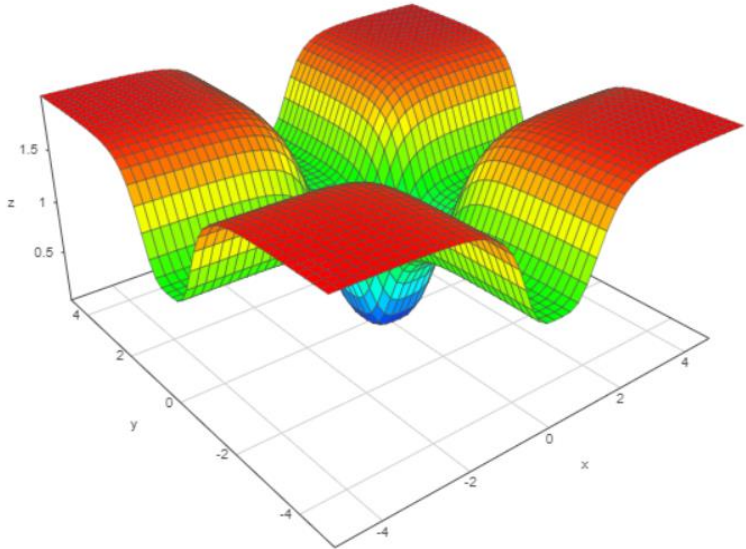

Figure 2. The Pagie function.

To automatically adapt the mutation rate to the characteristics of the problem and to the specific evolutionary phase we used the one-fifth success rule that vary the mutation rate so to maintain the proportion between offspring that are equal/better or worse than the parent to 1/5 (Rechenberg, 1973; Schumer and Steiglitz, 1968). This is realized by varying the mutation rate after the evaluation of each offspring on the basis of the following equation:

$$Mrate = \begin{cases} Mrate * 1.4 & \text{if } f(offsp.) \geq f(parent) \\ Mrate * 1.4^{-\frac{1}{4}} & \text{otherwise} \end{cases} \tag{3}$$

We will refer to the variants of the two algorithms described above that include the automatic adaptation of the mutation rate with the name (1 + 4) ES-AM and (1 + 4) ES-PL-AM.

The ratio between offspring that have a fitness equal or greater than their parent and offspring that have a lower fitness depends on the mutation rate, i.e. higher the mutation rate is higher the probability that the offspring receive maladaptive mutations is, and on the following three factors: (i) the local fitness surface, i.e. the greater the magnitude of the derivative of the local fitness surface, the greater the chance that mutations cause maladaptive effects is, (ii) the number of functional genes, the greater the number of functional genes, the greater the probability that mutations have maladaptive effect is, and (iii) the robustness to mutations, the greater the robustness of the current parent to mutations, the lower the probability that mutations cause a fitness loss is. The one-fifth success rule permits to appropriately tune the mutation rate to the variations of these factors occurring across generations.

## 5. Results

### 5.1 The $(1 + \lambda)$ ES-PL method largely outperforms the standard $(1 + \lambda)$ ES method

Table 1 reports the results of 18 series of experiments carried with the $(1 + \lambda)$ ES method in which we systematically varied the mutation rate (MutRate) and the number of offspring ($\lambda$). As can be seen, the best performance is achieved by setting $\lambda$ to 4 and MutRate to 2%.

|  | $\lambda = 1$ | $\lambda = 4$ | $\lambda = 7$ | $\lambda = 9$ | $\lambda = 19$ | $\lambda = 49$ |
|---|---|---|---|---|---|---|
| MutRate = 1% | 40% (321'753) | 70% (311'460) | 50% (516'851) | 33% (580'597) | 43% (562'594) | 30% (873'761) |
| MutRate = 2% | 50% (257'520) | **80% (274'560)** | 60% (437'930) | 40% (500'527) | 50% (471'340) | 30% (753'370) |
| MutRate = 4% | 40% (327'595) | 73% (307'512) | 60% (492'921) | 36% (609'512) | 36% (591'203) | 23% (980'844) |

Table 1 Performance of the best circuits evolved with the $(1 + \lambda)$ ES obtained by systematically varying the mutation rate and $\lambda$. The first number of each cell indicate the percentage of replications that successfully solved the problem over 30 replications. The numbers in parenthesis indicate the average number of evaluations required to find an optimal solution in the case of the successful replications.

The (1 + 4) ES-PL method, introduced in this paper, outperforms the standard (1 + 4) ES method both in term of performance (it solves the problem in all replications instead than in 80% of the replications) and in terms of speed (it requires 38'320 generations, on the average, instead than 59'746 generations, see Figure 3, Mann-Whitney U Test, p-value $< 10^{-4}$). In the case of (1 + 4) ES the average number of generations required to find an optimal solution is calculated on the 24 out of 30 replications that achieved optimal performance.

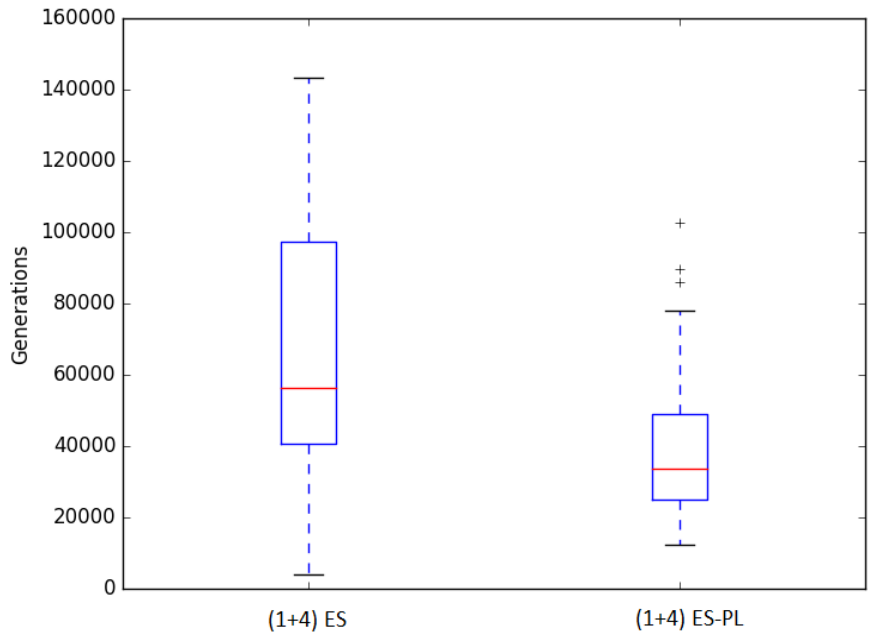

Figure 3. Number of generations required to solve the problem for experiments conducted with the (1 + 4) ES and with the (1 + 4) ES-PL with the best parameters ($\lambda = 4$ and mutation rate $= 2\%$). Data of the 24/30 successful replications carried with the (1 + 4) ES method and of the 30/30 successful replications carried with the (1 + 4) ES-PL method. Boxes represent the inter-quartile range of the data and horizontal lines inside the boxes mark the median values. The whiskers extend to the most extreme data points within 1.5 times the inter-quartile range from the box. "+" symbols indicate the outliers.

The programs evolved with the (1 + 4) ES-PL method are significantly larger than those evolved with the standard method both with respect to the number of functional nodes and with respect to the number of connections among nodes (Figure 4, Mann-Whitney U Test, p-value $< 10^{-4}$). Overall those data confirm that the preferential selection of functionally larger programs favors the evolution of larger solutions and boost evolutionary progress.

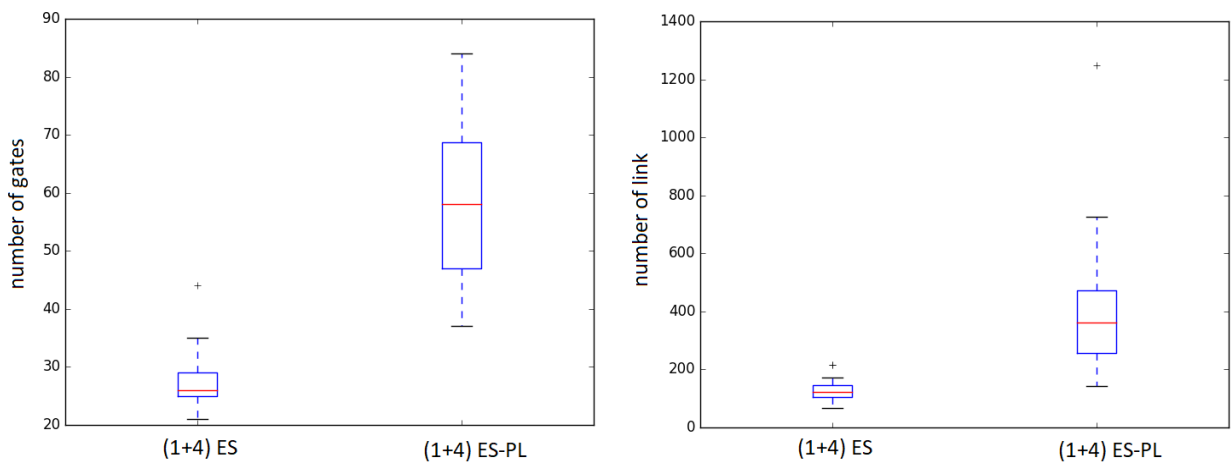

Figure 4. Size of the 24/30 successful programs evolved with the (1 + 4) ES algorithm and of the 30/30 successful programs evolved with the (1 + 4) ES-PL algorithm. The left figure shows the number of gates

forming the functional circuit. The right picture shows the number of links among gates of the functional circuits. Boxes represent the inter-quartile range of the data and horizontal lines inside the boxes mark the median values. The whiskers extend to the most extreme data points within 1.5 times the inter-quartile range from the box. "+" symbols indicate the outliers.

The program evolved with the $(1+4)$ ES algorithm are more robust to mutations than the programs evolved with $(1+4)$ ES-PL algorithm (Figure 5, left). On the other hand, programs evolved with the with $(1+4)$ ES-PL algorithm are more robust with respect to mutations affecting functional genes (Figure 5, right). This confirms that the program evolved with the former algorithm achieve robustness simply by minimizing the number of functional genes and consequently by minimizing the change that mutations affect functional genes. The programs evolved with the latter method instead achieve robustness through redundancy or degeneracy (Tononi, Sporns and Edelman, 1999; Edelman and Gally, 2001), i.e. through mechanisms that compensate the effect of mutations affecting functional genes.

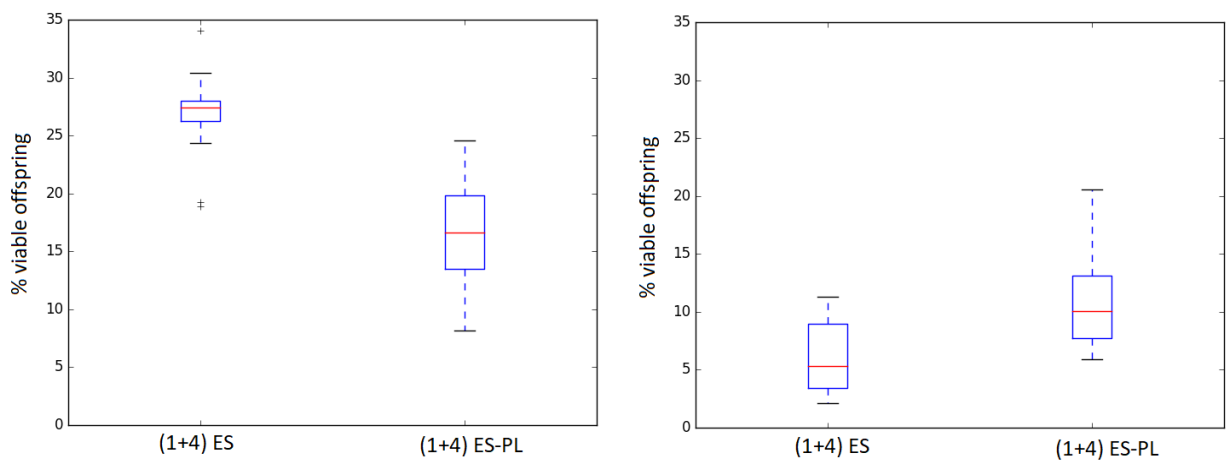

Figure 5. Left: Probability that offspring of circuits evolved with the two algorithms maintain the same fitness of their parents. Right: Probability that offspring of circuits evolved with the two algorithms with varied functional circuit maintain the same fitness of their parent. Data obtained by analyzing 100'000 offspring generated by the evolved circuit that achieved optimal performance. Boxes represent the inter-quartile range of the data and horizontal lines inside the boxes mark the median values. The whiskers extend to the most extreme data points within 1.5 times the inter-quartile range from the box. "+" symbols indicate the outliers.

## 5.2 The $(1+\lambda)$ ES-PL adapt faster to variations of the problem than the standard $(1+\lambda)$ ES method

In this section we show the results of a series of experiment carried out with the dynamically varying problem in which the category of a fraction of patterns vary every 100'000 generations.

Figure 6 show the average number of generations required by programs evolved with the two methods to adapt to variations of the problems. Both methods manage to find an optimal solution for the first and for the varied versions of the problem within 100'000 generations, in all cases. As shown by the figure, however, the programs evolved with the $(1+\lambda)$ ES-PL method required remarkably less generations to adapt to the variations of the problem than the programs evolved with the $(1+\lambda)$

ES method, independently from the number of patterns varied (Mann-Whitney U Test, p-value < $10^{-4}$ in all cases)

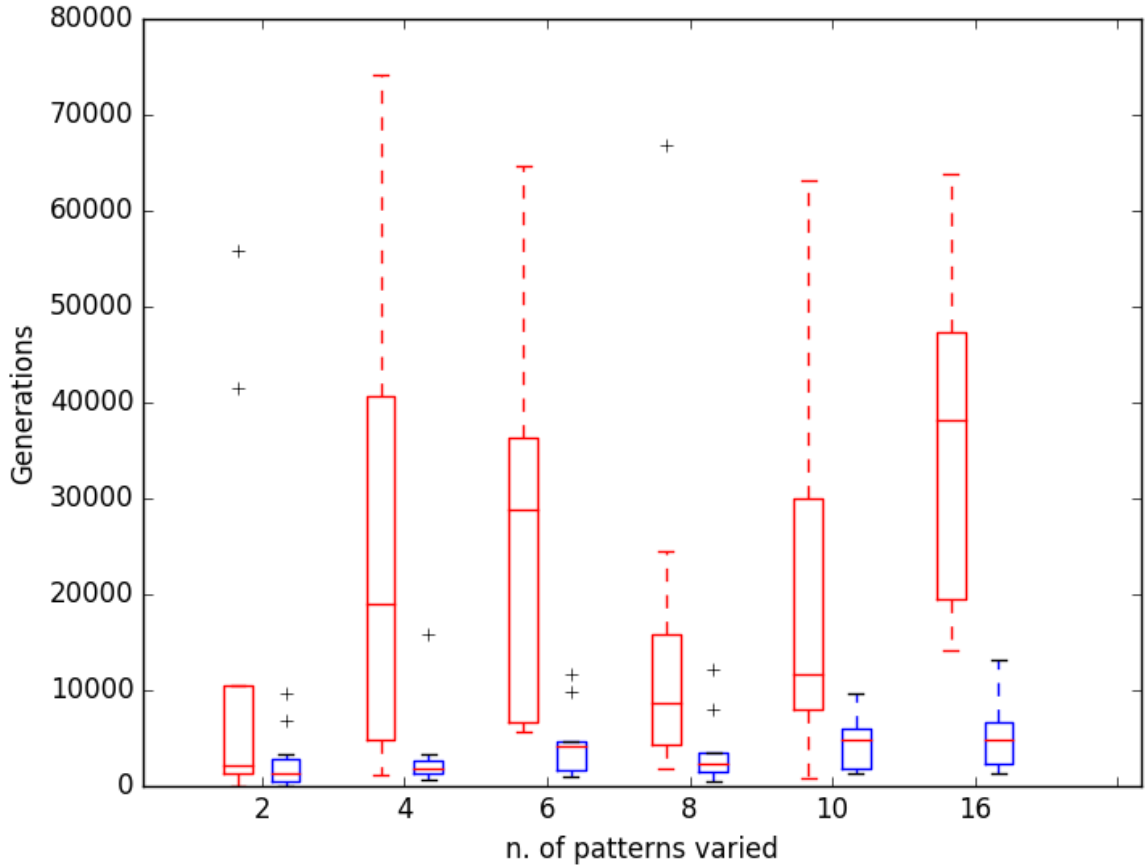

Figure 6. Average number of generations required by evolving circuits to adapt to a variation of the problem. The red and blue histograms indicate the data for the circuits evolved with the $(1 + \lambda)$ ES and the $(1 + \lambda)$ SO-ES algorithms respectively. The six pair of histograms indicate the results obtained in 6 corresponding experiments in which the desired output of 2, 4, 6, 8, 10, or 16 patterns selected randomly out of 32 total pattern was varied (i.e. switched from 0 to 1 or vice versa) every 100,000 generations. Boxes represent the inter-quartile range of the data and horizontal lines inside the boxes mark the median values. The whiskers extend to the most extreme data points within 1.5 times the inter-quartile range from the box. Circles indicate the outliers.

**5.3 Self-regulation of the mutation rate further boost performance**

To verify whether the preferential selection of larger solutions provide an advantage also on more complex versions of the problems and to verify whether the self-adaptation of the mutation rate further boost performance, we carried out additional experiments on a six, seven and eight-bit even parity problems with four algorithms that vary with respect to the usage of preferential selection of larger solution (PL) and with respect the utilization of self-adaptive mutation (AM).

The comparison of the results shown in Table 2 indicate that the (1+4) ES-PL method with preferential selection outperforms the (1+4) ES method also on the seven and eight-bit parity problems (Table 2, Mann-Whitney U Test, p-value < $10^{-4}$).

Moreover, the results indicate that the (1+ 4) ES-PL-AM method with adaptive mutation outperform the (1+4) ES-PL (Table 2, Mann-Whitney U Test, p-value < $10^{-4}$ ).

|  | (1+4) ES | (1+4) ES-AM | (1+4) ES-PL | (1+ 4) ES-PL-AM |
| --- | --- | --- | --- | --- |
| six-bit parity | 80% [274'120] | 83% [295'450] | 96% [210'987] | 100% [211'322] |
| seven-bit parity | 43% [417'822] | 45% [574'320] | 53% [379'313] | 90% [428'200] |
| eight-bit parity | 20% [725'532] | 16% [887'000] | 20% [625'482] | 76% [752'000] |

Table 2. Fraction of replications that achieved optimal performance on the six, seven and eight-bit even parity problem with the four considered algorithms. The numbers in parenthesis indicate the mean number of generations required to find optimal solution, in the replications that achieved optimal performance. In the case of the (1+4) ES and (1+4) ES-PL methods the mutation rate was set to 2%. In the case of the (1+4) ES-AM and (1+4) ES-PL-AM methods the mutation rate was initially set to 2% and then self-tuned. Each experiment has been replicated 30 times.

**5.4 Preferential selection of large solutions is beneficial also in continuous domains**

Finally, in this section we report the results obtained on the Paige regression problem. As shown in Table 3, also in this case the (1+4) ES-PL method outperforms the (1+4) ES method (Mann-Whitney U Test, p-value < 0.05) and the (1+ 4) ES-PL-AM method outperforms the (1+4) ES-PL method (Table 3, Mann-Whitney U Test, p-value < 0.05).

| (1+4) ES | (1+4) ES-AM | (1+4) ES-PL | (1+ 4) ES-PL-AM |
| --- | --- | --- | --- |
| 0% (66.96) | 3% (65.40) | 6% (59.81) | 16% (57.17) |

Table 3. Fraction of replications that achieved optimal performance on the Paige Function symbolic regression problem with the four considered algorithms. Performance are considered optimal when the given error is < $10^{-4}$. The numbers in brackets indicate the average error. In the case of the (1+4) ES and (1+4) ES-PL methods the mutation rate was set to 3%. In the case of the (1+4) ES-AM and (1+4) ES-PL-AM methods the mutation rate was initially set to 3% and then self-tuned. Each experiment has been replicated 30 times.

The advantage provided by the usage of preferential selection and adaptive mutation is much smaller in this problem than in the problems considered in the previous sections.

To increase the role of the preferential selection for larger solutions we decided to implement an additional variation of the algorithms that preferentially selected larger solutions also in the presence of quasi neutrality, i.e. also among solutions that do not have the same fitness but that differ in their fitness for less than 10%. We refer to these new algorithms with the term (1+4) ES-PLQS and (1+ 4) ES-PLQS-AM where QS stands for quasi-neutrality.

As shown in Table 4, the preferential selection of larger solutions also in the presence of quasi-neutrality provides a much greater advantage with respect to the preferential selection of larger solutions in the presence of full neutrality (Mann-Whitney U Test, p-value <0.01).

| (1+4) ES-PLQS | (1+ 4) ES-PLQS-AM |
| --- | --- |
| 26% (33.21) | 46% (25.16) |

Table 4. Fraction of replications that achieved optimal performance on the Paige Function symbolic regression problem with the (1+4) ES-PLQS and (1+ 4) ES-PLQS-AM algorithms. Performance are considered optimal when the difference between the error is < $10^{-4}$. The numbers in brackets indicate the average error. In the case of the (1+4) ES-PLQS methods the mutation rate was set to 3%. In the case of the (1+4) ES-PLQS-AM method the mutation rate was initially set to 3% and then self-tuned. Each experiment has been replicated 30 times.

# 6. Conclusions

In this paper we demonstrated how the efficiency of Cartesian Genetic Programming methods can be scaled up through the preferential selection of phenotypically larger solutions, i.e. through the preferential selection of larger solutions among equally good or similarly good alternative solutions.

The rationale behind the method proposed is that evolution tends to select solutions that are robust with respect to mutations and that the simplest way to achieve robustness consists in selecting programs that use a minimal number of nodes to produce their outputs. Indeed, smaller the number of functional nodes is, smaller the probability than these nodes are altered by mutations is. As pointed out in previous research (Raman and Wagner, 2011; Milano, Pagliuca and Nolfi, 2017), the chance to produce phenotypically varied programs as a result of genetic variations is positively correlated with the size of the solutions. Consequently, the tendency to select minimal solutions that are very robust to mutations reduces the probability to generate better solutions is successive generations which, in turn, reduces the efficacy of the evolutionary process. The preferential selection of larger solutions introduced in this paper enables to neutralize this negative effect and promotes the selection solutions that have a greater chance to generate better solutions as a result of genetic variation.

The advantage of the preferential selection of larger solutions has been validated on the six, seven and eight-bit parity problems, on a dynamically changing problem involving the classification of binary patterns, and on the "Paige" regression problem. In all cases, the preferential selection of larger solutions provides an advantage in term of the performance of the evolved solutions and in term of speed, i.e. number of evaluations required to evolve optimal or high-quality solutions. In the case of the dynamic changing problem, the preferential selection of larger solutions also enables to remarkably reduce the number of generations required to adapt the current solutions to variations of the problem.

The advantage provided by the preferential selection of larger solutions can be further extended by self-adapting the mutation rate through the one-fifth success rule (Rechenberg, 1973; Schumer and Steiglitz, 1968). The advantage of the combined effect of the preferential selection of larger solutions and of self-adaptation of the mutation rate is due to the fact that the robustness of evolving solutions is strongly influenced by their functional size for the reason explained above. Consequently, the dynamic regulation of the mutation rate becomes especially important when the functional size of the evolving solutions vary remarkably among generations.

Finally, for problems like the Paige regression in which neutrality plays a minor role, the advantage of the preferential selection of larger solutions can be further extended by preferring larger solutions also among quasi-neutral alternative candidate solutions, i.e. also among solutions achieving similar performance.

**Acknowledgements**

The authors thanks Prof. Julian Miller for insightful discussions and suggestions.